%% file: main_diffMeshing.tex
\documentclass[acmtog]{acmart}

\usepackage{booktabs} 
\usepackage{subcaption} 
\usepackage{tabularx} 

\usepackage[ruled]{algorithm2e} 

\SetAlFnt{\small}
\SetAlCapFnt{\small}
\SetAlCapNameFnt{\small}
\SetAlCapHSkip{0pt}

\newcommand{\dt}{\mathbf{DT}}
\newcommand{\wdt}{\mathbf{WDT}}
\newcommand{\dwdt}{\mathbf{dWDT}}
\renewcommand\footnotetextcopyrightpermission[1]{}

\acmDOI{}
\setcopyright{none}
\acmPrice{}
\acmISBN{}

\newcommand{\change}[1]{\textcolor{black}{#1}}
\begin{document}

\title{Differentiable Surface Triangulation}

\author{Marie-Julie Rakotosaona}
\affiliation{%
 \institution{LIX, École Polytechnique}
 \country{France}}
\email{mrakotos@lix.polytechnique.fr}

\author{Noam Aigerman}
\affiliation{
 \institution{Adobe Research}
 \country{USA}
}
\author{Niloy J. Mitra}
\affiliation{
 \institution{Adobe Research, University College London (UCL)}
 \country{United Kingdom}
}
\author{Maks Ovsjanikov}
\affiliation{
 \institution{LIX, École Polytechnique}
 \country{France}
}
\author{Paul Guerrero}
\affiliation{
 \institution{Adobe Research}
 \country{United Kingdom}
}

\renewcommand\shortauthors{Rakotosaona, M-J. et al}

\input{sections/abstract.tex}

%
%


%
%


\begin{teaserfigure}
  \includegraphics[width=\textwidth]{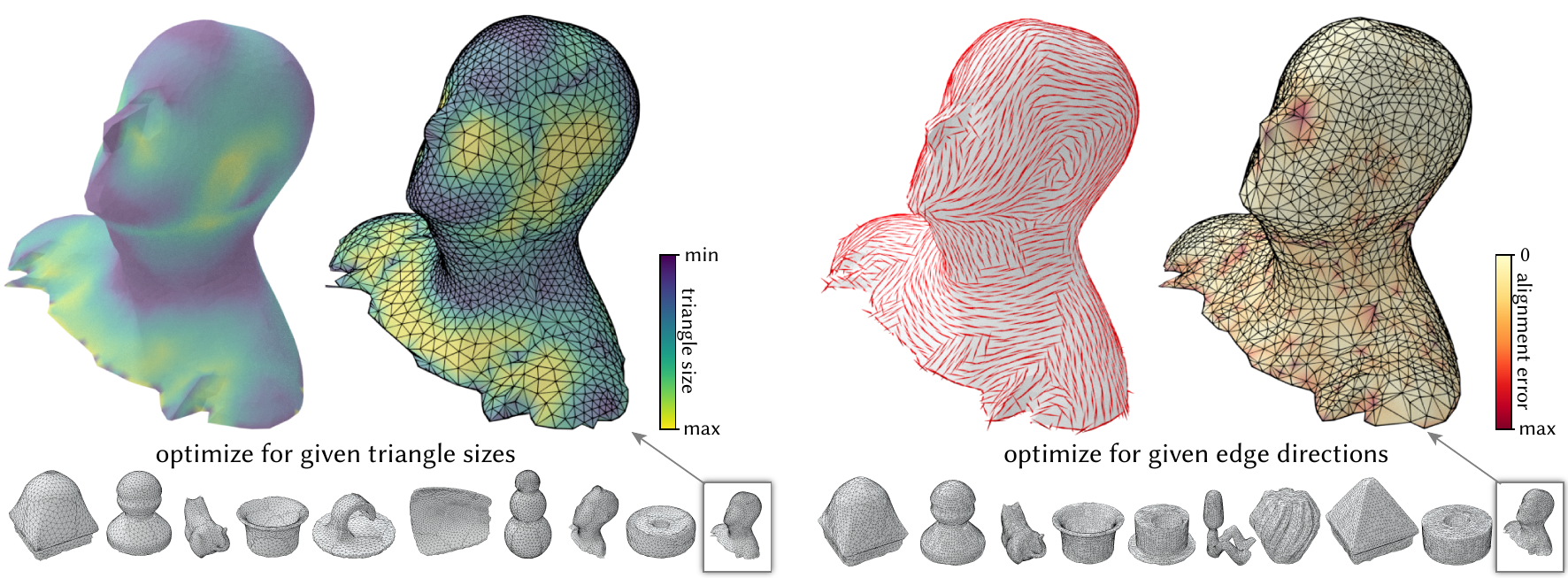}
  \caption{We present a fully differentiable approach for optimizing triangle meshes both in 2D and on surfaces. Our approach allows to optimize the mesh using any differentiable objective function, based on vertex positions or shapes of triangles using continuous optimization techniques. Here we demonstrate  meshes obtained on a surface by optimizing for (left) sizes of triangles to depend inversely on the mean absolute curvature value, and (right) alignment of triangle edges to the maximal principal curvature directions \change{(i.e. triangle edges tend to follow the vector field)}. This optimization is done in a fully differentiable manner without any post-processing or combinatorial operations such as edge flips or vertex splits. Our framework is general and can be thus integrated within modern optimization and learning modules. }
  \label{fig:teaser}
\end{teaserfigure}
\maketitle
\thispagestyle{plain}
\pagestyle{plain}

\input{sections/introduction.tex}

\input{sections/related_work.tex}

\input{sections/method.tex}

\input{sections/results.tex}

\input{sections/conclusion.tex}

\bibliographystyle{ACM-Reference-Format}
\bibliography{main_diffMeshing}

\end{document}

%% file: sections/abstract.tex
\begin{abstract}

Triangle meshes remain the most popular data representation for surface geometry. This ubiquitous representation is essentially a hybrid one that decouples continuous \emph{vertex locations} from the discrete topological \emph{triangulation}.
Unfortunately, the combinatorial nature of the triangulation prevents taking derivatives over the space of possible meshings of any given surface.
As a result, to date, mesh processing and optimization techniques have been unable to truly take advantage of modular gradient descent components of modern optimization frameworks.
In this work, we present a \textit{differentiable surface triangulation} that enables optimization for any per-vertex or per-face differentiable objective function over the space of underlying surface triangulations.
Our method builds on the result that \textit{any} 2D triangulation can be achieved by a suitably perturbed weighted Delaunay triangulation.
We translate this result into a computational algorithm by proposing a \change{soft} relaxation of the classical weighted Delaunay triangulation and optimizing over vertex weights and vertex locations.
We extend the algorithm to 3D by decomposing shapes into developable sets and differentiably meshing each set with suitable boundary constraints.
We demonstrate the efficacy of our method on various planar and surface meshes on a range of difficult-to-optimize objective functions. Our code can be found online: \href{https://github.com/mrakotosaon/diff-surface-triangulation}{https://github.com/mrakotosaon/diff-surface-triangulation}.
\end{abstract}

%% file: sections/introduction.tex
\section{Introduction}

Triangle meshes are arguably the most predominant surface representation, both in geometry processing and computer graphics, as well as in other fields such as computational geometry and topology. The popularity of triangle meshes comes from their simplicity, flexibility, and the existence of many data structures for efficient mesh navigation and  manipulation \cite{devillers2001walking,devillers2002delaunay,boissonnat2000triangulations,toth2017handbook}. Many methods have been developed to compute or modify triangulations of given surfaces or point clouds, while promoting properties such as alignment to shape features (e.g., ridges or creases), adapting sampling density to geometric detail, or triangle aspect ratio (see \cite{cazals2004delaunay,berger2017survey} for an overview).

Unfortunately, as of now, no method has been proposed to enable a continuous, differentiable representation of triangulations. This is mainly due to the fact that in addition to the \emph{continuous} spatial aspect - the position of each vertex - triangulations also have a   \textit{discrete} combinatorial component -  the connectivity, i.e., the set of edges and triangles connecting the vertices.  As a result, existing algorithms either optimize the mesh quality by moving the vertex locations while keeping their connectivity fixed \cite{nealen2006laplacian}, re-mesh from scratch, or iterate between updating the vertex positions and their connectivity, e.g., \cite{hoppe1993mesh,tournois2008interleaving}.

This lack of a unified differentiable representation is particularly unfortunate in light of recently-introduced  gradient-based optimization frameworks such as Pytorch~\cite{paszke2019pytorch} and TensorFlow~\cite{abadi2016tensorflow} for Machine Learning applications. These frameworks rely on the  differentiablity of the pipeline and enable modular design.
In absence of such a differentiable triangulation framework, current deep-learning pipelines either perform surface meshing during post-processing, or use formulations that are learned via proxies \cite{liao2018deep,sharp2020pointtrinet,liu2020meshing,rakotosaona2020learning}, which typically do not give explicit access to the resulting triangle mesh structure.

In this work, we devise what we believe to be the first  formulation for \textit{differential triangulation}, enabling gradient-based optimization for per-face and/or per-vertex objectives, such as size and curvature alignment. Our approach is general, can be applied to manifolds represented in any explicit representation, is modular, and supports optimizing for any objective that can be expressed as a differentiable function with respect to triangle properties like size and angles.

The main technical challenge in devising a differentiable triangulation is developing a smooth representation that allows to control both the vertex positions and the (inherently-combinatorial) mesh structure, while also ensuring the resulting mesh is always a 2-manifold. Our core idea is to use the concept of a weighted Delaunay triangulation~($\wdt$)~\cite{compgeom:2000}. It considers a given set of vertices, along with per-vertex weights, which define a unique triangulation using a Voronoi-like partition of space.

In this paper we propose a \emph{differentiable} weighted Delaunay triangulation ($\dwdt$), by considering (arbitrary) triplets of vertices and whether they constitute a triangle in the triangulation defined by the weights and vertices. While in classic $\wdt$, this existence receives a binary value, we generalize that  definition by assigning \change{inclusion} scores to triangle membership, thus giving them a \emph{soft} association. We demonstrate that this relaxation provides a unified control over both the vertices and the mesh structure, and can be used to directly optimize any (differentiable) objective function defined on the triangles. \change{Intuitively, we define the triangle \change{inclusion scores} in terms of Voronoi diagram distances that represent how close a certain triangle is from inclusion into (or removal from) the triangulation. Represented as a continuous quantity, we can optimize triangle inclusion scores as a function of vertex positions and weights.} Importantly, Memari et al.~\shortcite{memari2011parametrization} showed that, in 2D,  \emph{any} triangulation can be represented through a perturbation of a $\wdt$, \change{in other words, \emph{any} triangulation can be reached by  adjusting vertex positions and weights, and then applying a $\wdt$.} 
Therefore, our approach is both differentiable and generic, allowing to accommodate a wide range of mesh structures.

To apply our relaxation  to 3D surfaces, we decompose the source into local patches, and then perform per-patch differentiable meshing with appropriate boundary constraints.
For example, in Figure~\ref{fig:teaser} we show triangulations obtained by
optimizing for different objective functions, given the same original underlying surface models.  The modular nature of our approach makes it easy to switch between target objective functions. Similarly, we can triangulate different surface representations (see Figure \ref{fig:analytic} for a triangulation of an analytic surface defined by a function).

We evaluate our method to produce 2D and 3D meshes optimized for a mix of target objective functions such as shape/size of triangles, and alignment to given vector fields, thereby highlighting that  our approach is both more flexible, and can accommodate for more diverse objectives than alternative approaches.
%

%% file: sections/related_work.tex
\section{Related Work}
\label{sec:related}


Surface remeshing and triangle mesh optimization are both extremely well-studied problems in computational geometry, computer graphics, and related fields. Below we review methods most closely related to ours,  and refer to recent surveys, including \cite{Khatamian:2016,book:DT_2016,alliez2008recent,khan2020surface} for a more in-depth discussion.

\paragraph*{Simplification-based approaches}
A common objective for surface remeshing is reducing the number of elements in the final mesh. As a result, especially early remeshing techniques, starting with the pioneering QEM approach \cite{garland1997surface}, often focused on preserving mesh quality during simplification (see  \cite{garland1997surface} for a survey of local methods). Such methods are typically based on edge-collapse operation followed by vertex position optimization, and have been extended both in terms of efficiency, e.g., \cite{hussain2009efficient,ozaki2015out}, the use of various metrics \cite{ng2014simplification} including feature preservation \cite{wei2010feature}, and even using spectral quantities \cite{lescoat2020spectral} during edge collapse. However, such approaches are essentially greedy and typically do not allow to optimize mesh properties based on general structural criteria.

\paragraph*{Local methods}
A related set of methods includes approaches based on local mesh modification while aiming to improve the overall mesh quality, e.g., \cite{hu2016error,dunyach2013adaptive,yue20073d}. In addition to edge collapse, these local operators include edge flipping, edge splitting,
and vertex translation. A prominent method in this category is real-time adaptive remeshing (RAR) \cite{dunyach2013adaptive}, which uses an adaptive sizing function and edge flipping to optimize the mesh quality and vertex valence. This framework was recently extended for efficient \emph{error-bounded} remeshing \cite{cheng2019practical} through a use of a range of powerful local refinement operations. Similarly, Explicit Surface Remeshing (ESR) \cite{surazhsky2003explicit} is another efficient method for remeshing based on local refinement operations coupled with angle-based smoothing. The more recent Instant Meshes~\cite{jakob2015instant} technique advocates using local optimization and smoothing, while aiming to optimize potentially global consistency. This results in a powerful and efficient framework, capable of handing both isotropic triangular or quad-dominant meshes. Nevertheless, as with other local techniques the topology \change{(i.e. the connectivity between vertices)} and geometry are handled separately, preventing a unified differentiable, global mesh optimization.

\begin{figure*}[t!]
    \centering
    \includegraphics[width=\textwidth]{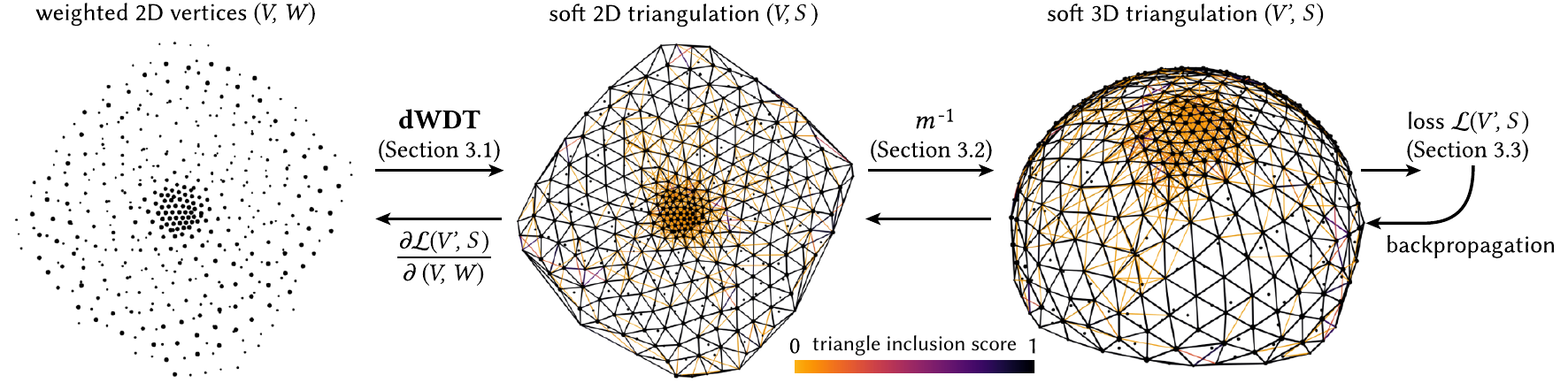}
    
    \vspace{-1mm}
    \caption{\textbf{Overview of our approach.} We propose a differentiable weighted Delaunay Triangulation ($\dwdt$) to create a soft triangulation from a set of 2D vertices $V$ with associated associated weights $W$ (shown as marker size). In the soft triangulation, triangles have \change{inclusion scores} $S$ of being part of the triangulation. We illustrate triangle \change{inclusion scores} as edge colors (using the largest \change{inclusion score} of the two adjacent faces) and only show triangles with \change{inclusion score} $> 0.001$. The 2D vertices $V$ are lifted to form a soft 3D triangulation on the manifold's surface using a fixed mapping $m^{-1}$. Since the pipeline is fully differentiable, we can propagate gradients of any differentiable loss on the 3D triangulation back to the vertex positions $V$ and weights $W$. Note that by choosing appropriate weights $W$, our network can ignore points and produce a triangulation over a subset of points, if desired.
\vspace{-1mm}}
    \label{fig:overview}
\end{figure*}

\vspace{-1mm}
\paragraph*{Delaunay and CVT-based methods}
Another powerful set of remeshing methods, more closely related to our approach are based on Delaunay triangulations, and centroidal Voronoi tessellations~(CVT). The former category includes approaches based on triangle refinement by flipping non-locally Delaunay (NLD) edges \cite{dyer2007delaunay} and defining an \emph{intrinsic} Delaunay triangulations \cite{fisher2007algorithm}. Furthermore, global optimization techniques have also been used for finding optimal Delaunay triangulations \cite{chen2011efficient} under the assumption that vertex connectivity is fixed. In a different line of work, centroidal Voronoi tessellations (CVT) have been used for finding an approximately uniform vertex distributions, so that their Voronoi diagram (and thus its dual, the Delaunay triangulation) is well-shaped, e.g., \cite{du2003constrained,yan2009isotropic,wang2015intrinsic} among many others. Such methods have also been extended, for example, to explicitly penalize obtuse and sharp angles \cite{yan2015non} and to \emph{anisotropic} remeshing by embedding in an appropriate (e.g., feature or curvature-aware) space \cite{levy2013variational}. Nevertheless, the final shape of the triangulation is difficult to control using these methods, and it is not easy to combine multiple objective functions in a coherent optimization strategy.

\vspace{-1mm}
\paragraph*{Optimization-based approaches}
Finally we also note methods based explicitly on optimizing an objective. This includes both local, e.g., \cite{hoppe1993mesh,dunyach2013adaptive}  and global optimization, e.g., \cite{valette2008generic,marchandise2014optimal}   strategies (see also Section 4.7 in \cite{khan2020surface}). Existing optimization strategies most often rely on either smoothness energies \cite{taubin1995signal,desbrun1999implicit,fu2014anisotropic}, use sampling~\cite{fu2009direct} or a variant of CVT, e.g., \cite{yan2014low} to optimize vertex positions. In both cases, while the \emph{positions} of the vertices can be optimized, the connectivity is only defined implicitly and updated separately, typically without explicitly taking into account the optimization objective. More fundamentally, the mesh structure is purely combinatorial, preventing the use of powerful tools based on differentiability.

In contrast to these approaches, we propose a fully differentiable framework that allows to jointly optimize for both vertex positions and triangle mesh connectivity, by using a soft version of the weighted Delaunay triangulation~($\wdt$). Our method is inspired by theoretical results  demonstrating that in 2D any triangulation can be represented through a perturbation of a $\wdt$~\cite{memari2011parametrization}. Importantly, the same result does not hold for the standard Delaunay triangulation, and therefore optimizing over the \emph{weights} of the $\wdt$ as well as the \textit{vertex positions} allows significantly more control over the shape of the final triangulation and even allows ignoring some input points if they are deemed unnecessary (i.e., high weights) in the final triangle mesh.

Importantly, the differentiable nature of our approach allows optimizing for a range of criteria jointly, simply by formulating a single (differentiable) objective function. Furthermore, it enables optimization of both vertex positions and criteria that depend on the connectivity in a unified framework. Finally, our differentiable meshing block can be also ultimately be used as part of a larger, differentiable shape processing or design system.

\vspace{-1mm}
\paragraph*{Weighted Voronoi and power diagrams.} Weighted Voronoi diagrams are also known as \emph{power diagrams}, and have been researched extensively in the context of triangulations \cite{glickenstein2005geometric}. In computer graphics, they have been used for various  tasks such as computing blue noise \cite{de2012blue}, or simulating fluid dynamics~\cite{de2015power}. \change{\cite{mullen2011hot,goes2014weighted,memari2011parametrization} considered power diagrams in the context of formulating different triangulation \emph{duals}. They also propose to optimize objectives on the dual. This is a different context and use-case than our differentiable formulation, which is geared towards a gradient-guided optimization of arbitrary geometric objectives on the \emph{triangulation}.}

\paragraph{\change{Differentiability in computer vision}} \change{Recently, with the success of deep learning in computer vision, making common operations differentiable has started to gain research interest. In particular relaxing hard condition for deep learning purposes into soft formulations has been used for  tasks such as RANSAC \cite{brachmann2017dsac}, rendering \cite{liu2019soft} or shape correspondence \cite{marin2020correspondence} among others. Similarly to these methods,  we present a soft formulation of triangle existence. }

%% file: sections/method.tex
\section{Method}
Let $(V', T)$ represent a triangulation of a surface in 3D space, with $V' = (v'_1, \dots, v'_n), v'_i \in \mathbb{R}^3$ vertices, and $T$ its triangular faces. A common strategy for triangulating a manifold surface is to first find a 2D parameterization that maps the surface to a planar 2D domain,  then sample a set of vertices $V = (v_1, \dots, v_n), v_i \in \mathbb{R}^2$ in the 2D domain, and compute a triangulation which respects the chosen vertices. Our method relies on the ubiquitous Delaunay triangulation~\cite{book:DT_2016, delaunay1934sphere} ($\dt$), used for triangulating a given 2D vertex set. We denote it as $T = \dt(V)$. A Delaunay triangulation always includes all chosen vertices, and is, uniquely defined with respect to them, as long as the points are in general position. In order to gain more control over the triangulation,  one can consider a \emph{weighted} Delaunay triangulation~\cite{toth2017handbook, aurenhammer1987power} $T = \wdt(V, W)$, where each vertex $v_i$ has a scalar weight $w_i$, with $W=(w_1, \dots, w_n)$. Traditional methods for computing a $\wdt$ are typically not differentiable, as the space of all possible faces is combinatorial.

We propose a differentiable weighted Delaunay triangulation $\dwdt(V, W)$ that is differentiable with respect to both the vertex positions $V$ and the weights $W$. In conjunction with a parameterization $m$ that defines a bijective and piecewise differentiable mapping $m$ from a surface in 3D to a 2D parameter space, $\dwdt$ enables a differentiable pipeline for triangulating 3D domains. We describe our differentiable triangulation approach in two parts (see Figure~\ref{fig:overview}). First, in Section~\ref{sec:diff_2d_meshing}, we describe the differentiable weighted 2D Delaunay triangulation $\dwdt$. In this part, we first focus on the definition of $\dt$ and the existence of triangles, w.r.t. vertex positions and weights, and then replace the  binary  triangle existence function with a smooth \emph{\change{triangle inclusion score}}, again defined w.r.t the vertex positions and weights, in a way that naturally follows from the definition of $\dt$. This yields a \emph{soft} and differentiable notion of a triangulation that can easily be generalized to a weighted Delaunay triangulation. Then, in Section~\ref{sec:3d_param}, we describe a parameterization $m$ that maps between a manifold surface and our 2D Delaunay triangulation to obtain $\dwdt$ on 3D surfaces, before describing the losses and optimization setup in Section~\ref{sec:losses}.



\subsection{Differentiable Weighted Delaunay Triangulation}
\label{sec:diff_2d_meshing}

Assume we are given a set of vertices $V=\{v_1,...,v_{|V|}\}$ with $v_j \in\mathbb{R}^2$. Consider the set of all possible triangles defined over these vertices, i.e., all possible triplets of vertices:
\begin{equation}
    T^* = \left\{\left(v_j,v_k,v_l\right) | v_j,v_k,v_l\in V \right\}.
\end{equation}
Any triangulation of the vertices $V$ is a subset of all possible triangles $T\subset T^*$ on $V$, and we can consider the triangulation's \emph{existence} function $e:T^* \to \left\{0,1\right\}$, defined for any triplet  $t_i\in T^*$ as
\begin{equation}
\label{eq:existence}
    e_i = \begin{cases}
      1 &t \in T\\
      0 &t \notin T.
    \end{cases}
\end{equation}
From this perspective, the binary and discrete existence function is the cause of the combinatorial nature of the triangulation problem. Hence, our main goal is to define a \emph{smooth} formulation in which this function is differentiable as to enable gradient-based optimization.
We achieve this by extending $\wdt$ to the smooth setting.

\begin{figure}[t!]
    \centering
    \includegraphics[width=\linewidth]{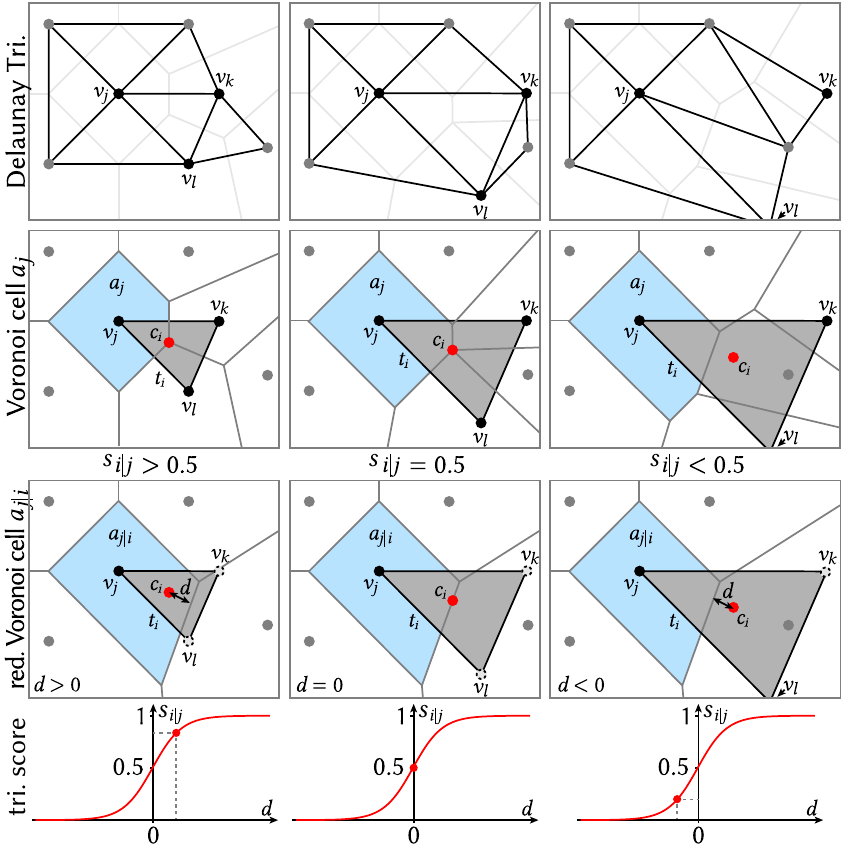}
    \caption{\textbf{Triangle \change{inclusion score} and reduced Voronoi cell.} A reduced Voronoi cell for a given triangle $t_i$ at a vertex $v_j$ is constructed from the point set that excludes the two other vertices of the triangle. A triangle $t_i$ exists in the Delaunay triangulation, as long as its circumcenter $c_i$ remains inside the reduced Voronoi cell. We base triangle \change{inclusion scores} $s_{i|j}$ on the signed distance from $c_i$ to the boundary of the reduced Voronoi cell.}
    \label{fig:voronoi}
\end{figure}

Towards gaining intuition into $\wdt$, let us first consider the classic, non-weighted \emph{Delaunay triangulation} $\dt(V)$ of a given set of vertices $V$. This triangulation is defined by considering each possible triangle $t \in T^*$ and deeming it as part of the triangulation $\dt(V)$ if and only if its circumcenter is the shared vertex of the three Voronoi cells centered at the triangle's vertices (see Figure~\ref{fig:voronoi} for an illustration). The Voronoi cell of vertex $v_j$ is defined as the set of points in $\mathbb{R}^2$ closer to $v_j$ than to any other vertex $v_k\in V$.

Said differently, each pair of vertices $(v_j, v_k)$ divides the 2D plane into two half-spaces: the set of points closer to $v_j$, denoted as $H_{j<k}$, and the set of points closer to $v_k$, denoted as $H_{k<j}$. The Voronoi cell $a_j$ centered at $v_j$ is defined as the intersection of half-spaces $a_j \coloneq \bigcap_{k\neq j} H_{j<k}$. The triangle circumcenter is the intersection point of the three half-space boundaries between the three vertex pairs that define its edges. Hence, we can define the existence function of the Delaunay triangulation for a triplet of vertices, $t_i = (v_j,v_k,v_l)$ with circumcenter $c_i$ as
\begin{equation}
\label{eq:hp_existence}
    e_i = \begin{cases}
      1 & \text{if}\ \ c_i = a_j \cap a_k \cap a_l \\
      0 & \text{otherwise}.
    \end{cases}
\end{equation}

\paragraph{Parameterizing Triangle Existence with respect to $V$. }
We are interested in how the triangulation $T$ changes as the vertex positions
are changed - namely, we aim to analyze the range of vertex positions
that do not change its membership function $e_i$ of a triangle $t_i$.

For any triangle $t_i = \left(v_j,v_k,v_l\right)$, we consider the three \emph{reduced} Voronoi cells $a_{j|i}$, $a_{k|i}$, $a_{l|i}$ respectively around the triangle's vertices $v_j$, $v_k$, $v_l$, where we define a reduced Voronoi cell $a_{j|i}$ centered at the triangle vertex $v_j$ as the Voronoi cell created by ignoring the two other vertices of the triangle, $v_k$ and $v_l$ (see Figure~\ref{fig:voronoi}). The triangle $t_i$ is part of the triangulation $T$ as long is its circumcenter $c_i$ remains inside the reduced Voronoi cells around its vertices. Similarly,  $t_i$ is not part of  $T$ as long as its circumcenter remains outside its three reduced Voronoi cells. Note that, by construction, the circumcenter simultaneously enters or exits the three reduced Voronoi cells. Thus, we can re-formulate the triangle existence $e_i$ as:
\begin{equation}
\label{eq:hp_existence_reduced}
    e_i = \begin{cases}
      1 & \text{if}\ \ c_i \in a_{x|i}\ \text{for any}\ x \in \{j, k, l\} \\
      0 & \text{otherwise}.
    \end{cases}
\end{equation}

\paragraph{Continuous Triangle \change{Inclusion Scores}}
We now turn to making $\dt$ differentiable by \emph{relaxing} the binary existence function $e_i$ defined in Equation \eqref{eq:hp_existence_reduced} into a continuous  \change{inclusion score} function, $s_i$, denoting the \change{inclusion score} a triangle $t_i \in T^*$ to exist as a member of the triangulation $T$, defined with respect to vertex positions. The \change{inclusion scores} are based on the signed distance of the triangle circumcenter to the boundary of the reduced Voronoi cells at the triangle vertices: considering a single vertex $v_j$ of the triangle $t_i$, and its reduced Voronoi cell $a_{j|i}$, the \change{inclusion score} is defined as:
\begin{equation}
\label{eq:prob}
    s_{i|j} := \sigma\big(\alpha\ d(c_i, a_{j|i}) \big),
\end{equation}
where $d$ is the signed distance (positive inside, negative outside) from a point $c_i$ to the boundary of a reduced Voronoi cell $a_{j|i}$, and $\alpha$ is a scaling factor for the width of the Sigmoid $\sigma$ (we use $\alpha = 1000$ in all experiments).
The Sigmoid gives a smooth transition from an \change{inclusion score} close to $1$ inside the reduced Voronoi cell to an \change{inclusion score} close to $0$ outside, with an \change{inclusion score} $0.5$ if the circumcenter lies on the boundary of the reduced Voronoi cell, i.e., exactly when the discrete triangle membership changes. The triangle \change{inclusion score} $s_i$ can then be defined as the average over the three \change{inclusion scores} at its vertices $v_j$, $v_k$, and $v_l$:
\begin{equation}
    s_i = \frac{1}{3} (s_{i|j} + s_{i|k} + s_{i|l}).
\end{equation}
Note that since the circumcenter simultaneously enters/exits the reduced Voronoi cells around each vertex, all three \change{inclusion scores} equal $0.5$ at a discrete membership transition.

For each triangle $t_i$, we store the triangle \change{inclusion score} $s_i$ and the three \change{inclusion scores} $s_{i|j}$, $s_{i|k}$ and $s_{i|l}$ defined for its three vertices, yielding a \emph{soft} 2D triangulation $(V, S)$ with \change{inclusion scores} $S$. We store the \change{inclusion scores} $s_{i|j}$ in addition to the triangle \change{inclusion scores} $s_i$, since most losses that we use are defined on vertices where using a triangle's vertex \change{inclusion scores} is more convenient. We can, subsequently, convert this soft triangulation into a discrete 2D triangulation $(V, T)$, by selecting all triangles where $s_i > 0.5$.
This gives us the same results as the discrete $\dt$, the final triangulation is guaranteed to be manifold.

Since the number of all possible triangles $T^*$ grows cubically with the vertex count, we reduce the number of triangles under consideration by observing that vertices in the triangles of a Delaunay triangulation are typically within the $k$-nearest neighbors of each other, for some small $k$ (we use $k=80$ in all experiments). Thus, at each Voronoi cell $a_j$, we only consider triangles that are within the $k$-nearest neighbors of $v_j$ and set all other triangle \change{inclusion score}
implicitly to $0$.
Note that since the 2D vertices $V$ change positions during optimization, we recompute nearest neighbours after each iteration of our algorithm.

\paragraph{Weighted Delaunay triangulation} Our relaxed formulation of the Delaunay triangulation can naturally be extended to the \emph{weighted} Delaunay triangulation $\wdt$, where weights are associated to each vertex.
The weights allow shifting the boundary between the two half-spaces $H_{j<k}$ and $H_{k<j}$, by the relative weights $w_j$ and $w_k$ of the two vertices. The weighted half space $H_{j<k}$ is defined as the set of points $x\in\mathbb{R}^2$ where
\begin{equation}
\label{eq:power_dist}
     \|x - v_j\|_2^2 - w_j^2 \leq \|x - v_k\|_2^2 - w_k^2
\end{equation}
so that a larger weight pushes the boundary away from the vertex. This
allows generalizing the definition of the Voronoi cell to a weighted Voronoi cell. As a result, the existence function Equation \eqref{eq:hp_existence_reduced} and the \change{inclusion score} $s_{i|j}$ in Equation~\eqref{eq:prob} can be used as-is, with the modified definition of the half-planes, and considering the \emph{weighted} circumcenter of the triangle. This makes the \change{inclusion scores} $S$ of the soft triangulation $(V,S)$ a function of both the vertex position $V$ and their weights $W$.

Thus, weights enable further control over the resulting triangulation, by enabling modifications the Voronoi cells (and therefore, the triangulation itself). In fact, note that it is even possible for a vertex to be excluded from a $\wdt$ (i.e., not be part of any triangle), if the weight difference to any other vertex is so large that the boundary line between the two half-spaces shifts past one of the vertices - a property not possible with classical Delaunay triangulation. We will make use of this property to allow our method to ignore vertices deemed unnecessary, hence producing triangulations with a reduced number of vertices. In the following, we consider the weighted triangle circumcenters, denoted by $c_i$, and the weighted Voronoi cells, denoted by $a_i$.



\subsection{3D Surface Parameterization}
\label{sec:3d_param}
So far we have defined differentiable triangulations of 2D sets of vertices. In order to
 apply our $\dwdt$
on a 3D surface $\mathcal{M}$, we reduce the problem to a set of 2D (triangulation) problems.

First, we construct a bijective piecewise differentiable mapping $m$ between the manifold and the 2D plane, i.e., a 2D parameterization. Next, we elaborate on the computation of this parameterization. 
As a pre-process, since we are not concerned with the original triangulation but only the underlying surface it represents, we initially remesh input models using isotropic explicit remeshing~\cite{cignoni2008meshlab} to yield meshes constituting between $3.5-4.5$K triangles. \change{We normalize each model to unit area.} We then decompose the manifold into a set of separate patches $\{\mathcal{P}_1, \mathcal{P}_2, \dots\}$ that can be individually parameterized with less distortion than the whole shape. Individual patches are found with a spectral clustering approach~\cite{ng2002spectral}, using the adjacency matrix for affinity. We used $10$ patches in all experiments.
%

Then, we construct a low-distortion mapping $m$ between the surface of a patch $\mathcal{P}_i$ and the 2D plane using Least-Squares Conformal Maps~\cite{Levy2002LSCM} (LSCM). To lower the distortion of the mapping for patches that are far from developable, we first measure the distortion as the deviation of the local scale factor from the global average.
Patches with high distortion are cut along the shortest geodesic between the area of maximum distortion and any existing boundary. This process is repeated until the maximum distortion of all patches, measured as the ratio between local scale and global average is above $15\%$ and the mapping is bijective. \change{Finally, we normalize the 2D parametrization of each patch to have equal average edge length.} 
We compute this mapping once, as a preprocess, and reuse it in all steps of the optimization.

\paragraph{Differentiable 3D Surface Triangulation}
Given the mapping $m$, we can pull back the computed 2D triangulation $(V, T)$ to a part of the 3D surface $\mathcal{P}_i$ using the inverse mapping $m^{-1}$.
Thus, our differentiable triangulation of a 3D surface patch is defined as:
\begin{equation}
(V', S) := \big(\big(m^{-1}(v_1), \dots, m^{-1}(v_n)\big),\ \dwdt(V, W)\big),
\end{equation}
which gives us the soft 3D triangulation $(V', S)$ that consists of a set of 3D vertices $V'$ and triangle \change{inclusion scores} $S$. Note that the \change{inclusion scores} are differentiable functions of the 2D vertices $V$ and their weights $W$, and that we can obtain a manifold discrete mesh at any time by selecting all triangles with \change{inclusion scores} $> 0.5$.
Since the mapping $m$ is piecewise differentiable, any loss $\mathcal{L}$ can be applied directly to the 3D vertices $V'$ and triangle \change{inclusion scores} $S$, allowing gradients to propagate back to the parameters $V$ and $W$ that define $V'$ and $S$. \change{We highlight that similarly to Leaky ReLU activations, the piecewise differentiability does not significantly impact optimization.}
%
%
We discuss the losses we use in our experiments in Section~\ref{sec:losses}.

\paragraph{Boundary preservation}
Special care must be taken to preserve the boundary of each patch, so that putting the patches back together does not result in gaps or overlaps. We use a two-part strategy to ensure pieces fit back together. First, we define a loss that repels vertices from the boundary of a patch, which we describe in Section~\ref{sec:losses}. Second, we perform a post-processing step that cuts the 2D mesh $(V, S)$ along the 2D boundary, based on a triangle flipping strategy along the boundary. \change{Namely, we use the simple strategy described in \cite{sharp2020you} between consecutive boundary points of the optimized patches. The boundary between patches is therefore kept fixed before and after the optimization step.}





\subsection{Losses and Optimization}
\label{sec:losses}
Our differentiable triangulation allows us to optimize a triangular mesh on a surface in 3D using any differentiable loss defined on the 3D vertex positions $V$ and triangle \change{inclusion scores} $S$. We experiment with several different losses, combinations of which are useful for both traditional applications, as well as novel ones, as we experimentally show in Section~\ref{sec:results}.

The \emph{triangle size loss} $\mathcal{L}_s$ encourages triangles to have a specified area:
\begin{equation}
    \mathcal{L}_s(V', S) := \frac{1}{\sum_{i,j} s_{i|j}} \sum_{i,j} s_{i|j}\ \left(0.5\ \|(v'_k-v'_j) \times (v'_l-v'_j)\|_2 - \change{A(v_j)} \right)^2,
\end{equation}
where $v'_j$, $v'_k$, and $v'_l$ are the 3D vertices of triangle $t_i$, \change{and $A(v_j)$ is the target area at vertex $v_j$, where $A$ is defined as a continuous function over the 3D surface.}
This loss allows us, for example, to coarsen a triangulation, when used in conjunction with other losses. Note that the size of the triangles is not constrained by the initial number of vertices - due to the $\wdt$ our optimized result can contain fewer vertices than the initial triangulation.

\begin{figure}[t!]
    \centering
    \includegraphics[width=\linewidth]{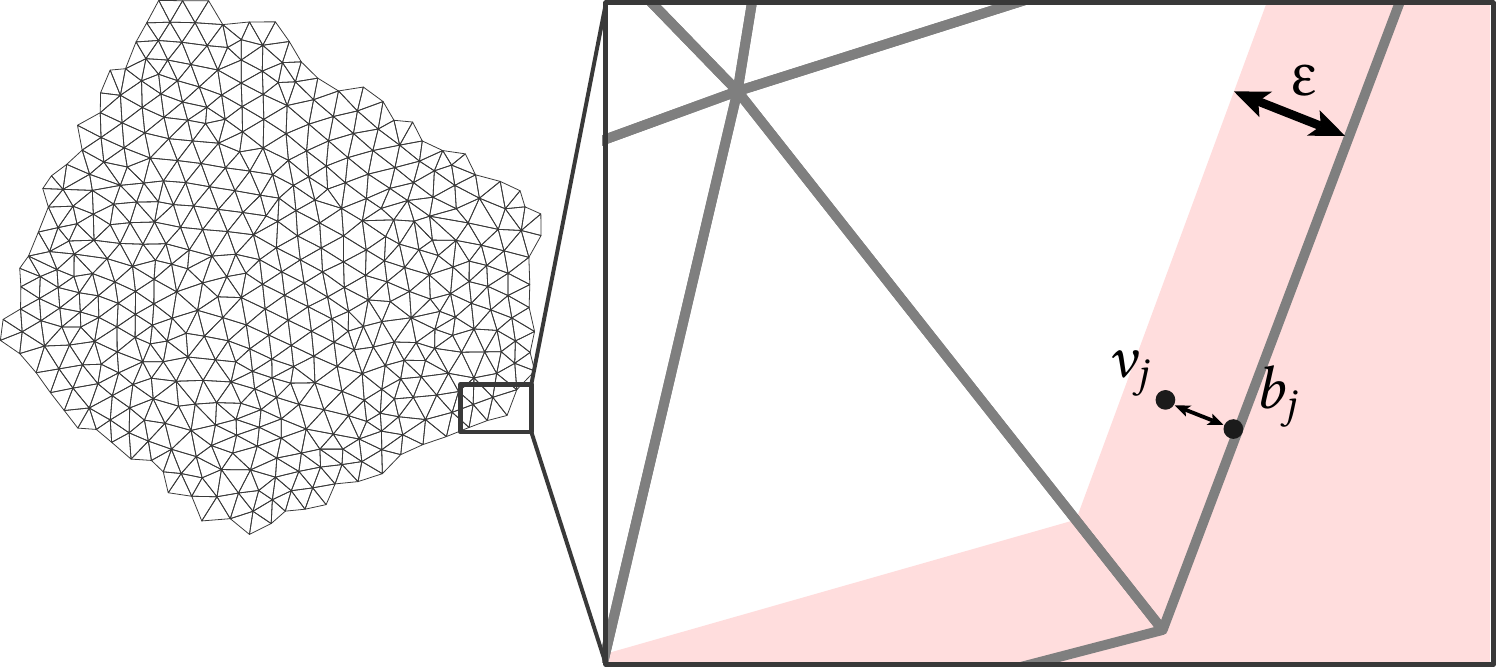}
    \caption{\change{\textbf{Boundary repulsion loss.} The repulsion loss is non-zero below a (signed) distance $\epsilon$ from the boundary. Non-boundary vertices inside the red region are pushed towards the center of the patch. }}
    \label{fig:repulsion_loss}
\end{figure}

The \emph{boundary repulsion loss} $\mathcal{L}_b$ encourages vertices to stay inside the 2D boundary of the patch $\mathcal{P}$ during the optimization:
%
\begin{equation}
    \mathcal{L}_b(V, \mathcal{P}) := \frac{1}{|V|} \sum_j  e^{\epsilon - \min\left(\epsilon,\ (v_j - b_j) n^b_j\right)}, 
\end{equation}
where $b_j$ is the point on the boundary closest to the vertex $v_j$ and $n^b_j$ is the 2D boundary normal at that point (pointing inward). The repulsion loss is non-zero below a (signed) distance $\epsilon$ from the boundary as we show in Figure \ref{fig:repulsion_loss}. We set $\epsilon$ to 0.01 in our experiments. Note that we do not use our \change{triangle inclusion scores} in this loss, since we want all vertices to remain inside the boundary, irrespective of \change{inclusion scores}. \change{We note that since   $\mathcal{L}_b$  has a local effect and does not rely on global properties of the patch, patches can be non-convex. }

The \emph{angle loss} $\mathcal{L}_a$ encourages triangles to be equilateral.
%
%
\begin{align}
\mathcal{L}_a(V', S, \mathcal{P}) := \frac{1}{\sum_{i,j} s_{i|j}} \sum_{i,j} s_{i|j} \ \left| \cos(\angle_j) - \cos({\pi}/{3}) \right|,
\end{align}
where $\angle_j$ is the corner angle of triangle $t_i$ including vertex $v_j$. Note that this loss can be modified to produce isosceles triangles.
%

The \emph{curvature alignment loss} $\mathcal{L}_c$ encourages two edges per vertex to align to the two directions of the minimum principal curvature vector field $C$. We define it as,
\begin{align}
\mathcal{L}_c(V', S, \mathcal{P}, C) :=& \frac{-1}{\sum_{i,j} s_{i|j}} \sum_j
    \Big( \nonumber\\
    & \text{LSE}(\cup_{i \in \mathcal{N}_j}\{\change{C(v_j)} \cdot h_{jk}\ s_{i|j}, \change{C(v_j)} \cdot h_{jl}\ s_{i|j} \}) \nonumber\\
    +\ & \text{LSE}(\cup_{i \in \mathcal{N}_j}\{-\change{C(v_j)} \cdot h_{jk}\ s_{i|j}, \change{-C(v_j)} \cdot h_{jl}\ s_{i|j} \}) \Big) \nonumber\\
    \text{with } h_{jm} =& (v'_j-v'_m)/{\|v'_j-v'_m \|_2},
\end{align}
where $\mathcal{N}_j$ are the triangles adjacent to $v'_j$, and $v'_j$, $v'_k$, $v'_l$ are the 3D vertices of triangle $t_i$. LSE denotes the smooth maximum function LogSumExp over the weighted alignment scores of all edges adjacent to vertex $v'_j$, where each triangle contributes two edges corresponding to $h_{jk}$ and $h_{jl}$. Intuitively, we want to maximize the alignment of the best-aligned edge in a star of each vertex, for both the positive and negative target guidance direction \change{$C(v_j)$, which is the principal curvature field evaluated at $v_j$.}

\paragraph{Optimization}

Given a loss $\mathcal{L}$, as a sum of a selection of the terms above,  we optimize the 3D mesh $M$, parameterized by the 2D vertex positions $V$ and vertex weights $W$. Since our framework is completely differentiable, we use the Adam~\cite{DBLP:journals/corr/KingmaB14} optimizer. We initialize all vertex weights with random values and use the mapping of the input mesh vertices to 2D as the initial 3D vertex positions.
%
We use a learning rate of $0.0001$ in all experiments. Please refer to the supplementary video for evolving triangulations over optimization iterations.

%% file: sections/results.tex

\begin{figure*}[t!]
    \centering
    \includegraphics[width=\textwidth]{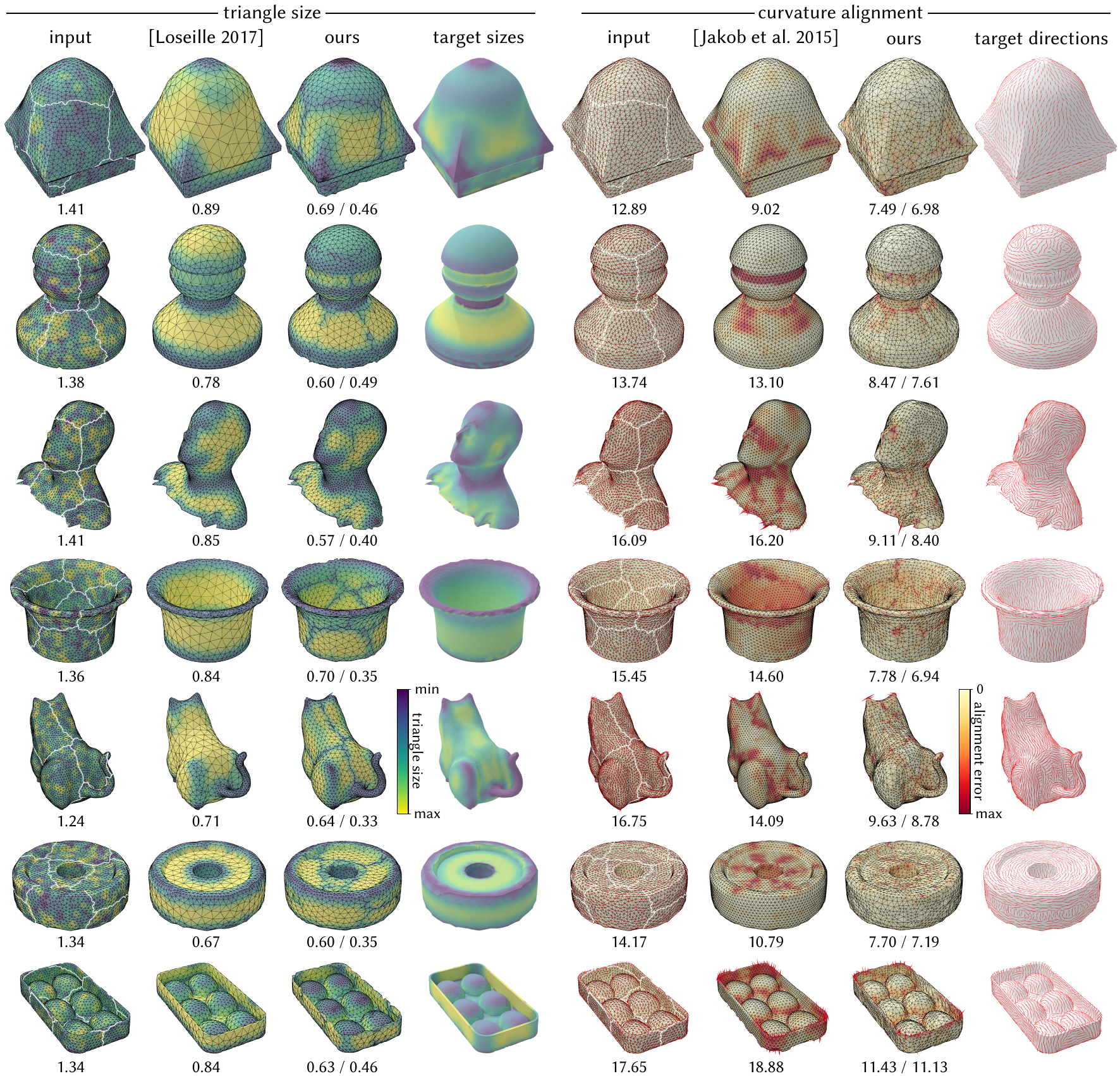}
    \caption{\textbf{Qualitative Results.} We show two applications of our approach. In the left half of the figure we optimize for given target triangle sizes, and compare with a state-of-the-art remeshing method~\cite{loseille2017unstructured} (triangles are colored according to size). In the right half, we optimize for edges that are aligned to the principal curvature directions and compare with Instant Meshes~\cite{jakob2015instant} (colors illustrate alignment errors). \change{Boundaries of the patch decomposition are shown as white lines on the input meshes.} The average error is given below each result - for our method we give the error with / without faces adjacent to a patch boundary. Note that our differentiable triangulation more accurately satisfies the target triangle sizes or edge directions.}
    \label{fig:results_qualitative}
\end{figure*}

\section{Results}
\label{sec:results}
We next describe experiments that highlight the key advantage of our method - differentiability, which enables plugging in and mixing any combination of differentiable losses, circumventing the need to design a specialized optimization method for each loss combination. 
Practically, the experiments show the efficacy of our method, and its ability to produce superior results than state-of-the-art methods that are specifically tailored to those specific applications. Code of our method is available at \url{anonymous.code}.

\subsection{Customized Triangulation}
Most triangulation tasks are formulated via user-provided requirements that are imposed on the resulting triangulation, such as desired triangle sizes or edge alignment. We employ our differentiable losses in two common scenarios, shown in Figures~\ref{fig:teaser} and~\ref{fig:results_qualitative}. We evaluate our method in both scenarios on $140$ randomly selected meshes (among those with genus $10$ or less) sampled from Thingi10k~\cite{zhou2016thingi10k}.

\paragraph{(i) Triangle size.} We first optimize the triangulation to match a given distribution of triangle sizes, represented as a scalar field over the surface. We chose to assign sizes that are the reciprocal of the mean absolute curvature value, so that high curvature regions receive a finer tessellation than lower-curvature regions. We sum the losses $\mathcal{L}_s$, $\mathcal{L}_b$, and  $\mathcal{L}_a$ with weights $0.5$, $500$, and  $10^7$, respectively, in order to scale each loss to the same range. Qualitative results are shown in the left half of Figure~\ref{fig:results_qualitative}.

As evaluation metric, we take the absolute difference between the resulting triangle size and the target size distribution. Since we are interested in the distribution of relative triangle sizes rather than the absolute sizes, we normalize the triangle sizes per model to have zero mean and unit standard deviation. To compare our triangle sizes to the continuous target size distribution, triangle size at each vertex is defined as the average size of all adjacent triangles. In Figure~\ref{fig:results_qualitative}, normalized triangle sizes are shown as colors  while the numbers below each result show the RMSE over all vertices.

We compare our method to the remeshing method of Loseille~\shortcite{loseille2017unstructured}, a state-of-the-art method for remeshing that can be guided by a given triangle size field, and show a qualitative comparison on a subset of shapes in Figure \ref{fig:results_qualitative}.
In most cases our method can reproduce the target size distribution more accurately. Note, for example, the size distribution on the top of the pawn, on the heads, on the rim of the hat and on the cat's hind.

\paragraph{(ii) Vector-field alignment.} In our second scenario, we optimize 3D meshes
with the loss $\mathcal{L}_c$ that encourages edges to align with a given vector field. We chose to use minimum principal curvature directions to encourage meshes which edges that adhere to ridge lines and geometric features. At the same time, we emphasize that any other user-prescribed field could be used as well. We minimize the loss $\mathcal{L}_c$ combined with the boundary repulsion loss $\mathcal{L}_b$ with weights of 1 and 500, respectively. Qualitative results are shown in the right half of Figure~\ref{fig:results_qualitative}.

\begin{figure*}[t!]
    \centering
    \includegraphics[width=\textwidth]{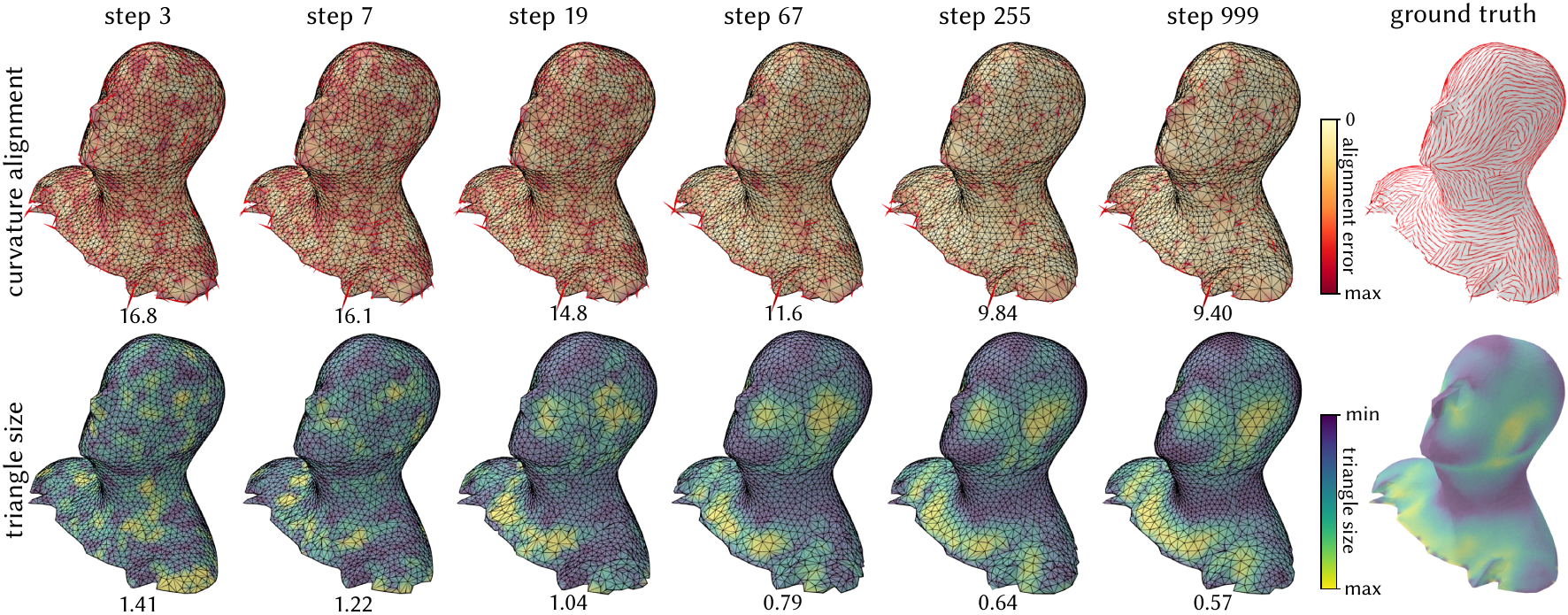}
    \caption{\textbf{Optimization steps.} We show our results at different optimization steps for curvature alignment (top row), and triangle size (bottom row).  }
    \label{fig:opt_steps}
\end{figure*}

As evaluation metric, we take the absolute angular difference between both the positive and negative prescribed curvature direction at each vertex and the best-aligned edge. We compare with Instant Meshes~\cite{jakob2015instant}, a method specialized to creating feature-aligned equilateral triangulations. Since Instant Meshes is designed to align to sharp features and is not well defined near flat regions or umbilical points, we weight the per vertex-alignment error using the following term: 
\begin{equation}
\label{eq:curvature_loss_weight}
w_j = \frac{|k_1^j-k_2^j|}{0.5*(|k_1^j|+|k_2^j|)},
\end{equation}
where $k_1^j$ and $k_2^j$ are the signed principal curvature magnitudes at vertex $j$. Intuitively, this term reduces the influence of regions that are nearly flat or umbilical, so as to not penalize the baseline in those regions unfairly. In Figure~\ref{fig:results_qualitative}, edge alignment errors are shown as colors while the numbers below each result show the RMSE over all vertices.


Our general-purpose triangulation achieves significantly better alignment, as can be seen by the significantly lower color-coded and average error on all models. While the baseline method of \cite{jakob2015instant} generates triangles that are very close to equilateral, the alignment with the curvature directions suffers, as can be seen on the lower part of the pawn, where none of the edges align well with the curvature directions. Similarly, for the cylindrical hat, our method generates edge-loops ``hugging'' the cylinder, while Instant Meshes does not present such edge-loops. On more organic models, such as the cat and human, lack of alignment is even more evident, e.g., on the human's brow. 

\paragraph{Quantitative evaluation} We further evaluate our method on our complete dataset of  $140$ meshes taken from Thingi10k~\cite{zhou2016thingi10k}. The quantitative results in Table~\ref{tab:trig_size} show the RMSE of the metrics described above over all vertices and all shapes in the dataset. Since the vertices at the boundary of our patches cannot fully be optimized with our approach, we provide errors computed both with and without the vertices at the patch boundaries.
In both cases and in both applications, our method approximates the correct triangle sizes and edges directions significantly better than the state-of-the-art methods~\cite{loseille2017unstructured} and~\cite{jakob2015instant}.

\begin{table}[h!]%
\caption{\textbf{Quantitative results.} We compare both the triangle size and curvature alignment applications to state of the art remeshing methods. For our results, we provide values computed both with and without the boundary triangles.}
\label{tab:trig_size}
\begin{minipage}{\columnwidth}
\begin{center}

\begin{tabularx}{\linewidth}{llll}
  \toprule
     input mesh & \cite{loseille2017unstructured}  & ours w/o bound. & ours\\ \midrule
     1.320& 0.865 & 0.499  & 0.686\\
  \bottomrule
\end{tabularx}
\subcaption*{triangle size}

\begin{tabularx}{\linewidth}{llll}
  \toprule
     input mesh & \cite{jakob2015instant} & ours w/o bound & ours  \\ \midrule
     14.043 &11.850 & 7.919 & 8.4617 \\
  \bottomrule
\end{tabularx}
\subcaption*{curvature alignment}
\end{center}
\end{minipage}
\vspace{-10pt}
\end{table}%

\subsection{Optimization}
\begin{figure*}
    \centering
    \includegraphics[width=\linewidth]{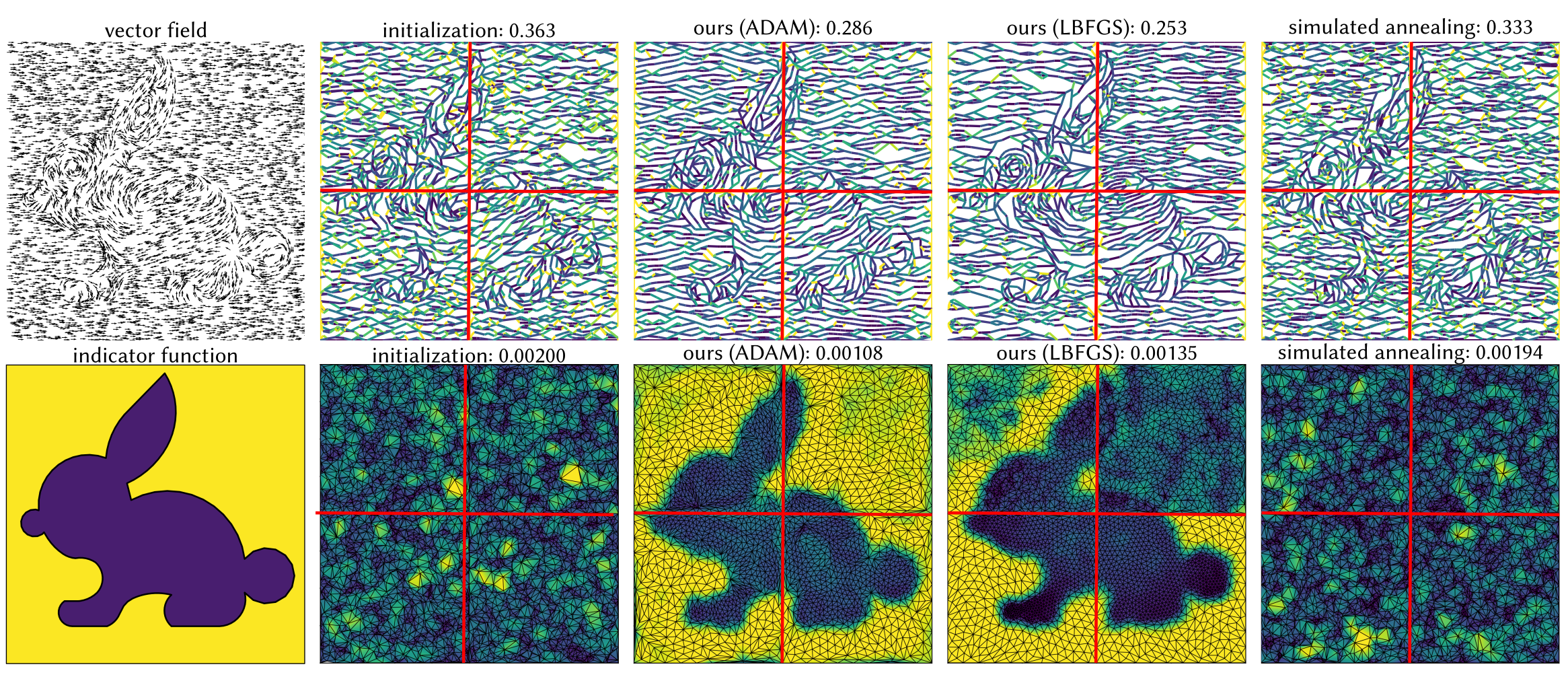}
    \caption{\textbf{Comparing different optimizers.} We compare ADAM, LBFGS and Simulated Annealing on a 2D mesh. We start from a 2D mesh with random vertices. In the top row, we optimize edges to align with a given vector field. The best-aligned edges are color-coded according to the alignment error (blue is lowest error, yellow largest error). The average alignment error is shown at the top. In the bottom row, we optimize triangle areas to align with a given size field. Vertices are color-coded according to the average neighboring triangle area (blue are smaller triangles, yellow larger triangles), with RMSE shown at the top. Note how the two gradient-based optimizers ADAM and LBFGS perform significantly better than the gradient-less simulated annealing.}
    \label{fig:optim_choice}
\end{figure*}

\paragraph{Choice of optimizer.} We evaluate the effect of different optimization methods, comparing ADAM, LBFGS, and Simulated Annealing. In Figure \ref{fig:optim_choice} we show results on a 2D triangulation example where we optimize for both triangle sizes, and alignment to a custom vector field. We run each optimizer for 1000 steps and observe that while LBFGS can achieve better performances on some patches, ADAM produces good results more consistently, and hence we opted to use it in all our experiments. \change{We use Simulated Annealing (SA) with the discrete mesh representation instead of our formulation as SA does not handle gradients. After computing the non differentiable weighted Delaunay triangulation,  we minimize the discrete version of our losses: for instance we align existing edges to the curvature vector field and fit the area of existing triangles to the target area function. } Both gradient-based methods perform significantly better than the non-gradient based method, Simulated Annealing, suggesting that our search space is typically too complex to allow for a more random search strategy that is not guided by gradients. We included the comparison to the non-gradient-based simulated annealing to show gradient-based methods are more apt for this problem; however, putting performance aside, we note that simulated annealing cannot accomplish the main goal of our work, which is to devise a triangulation module that can be used within differentiable optimization frameworks (e.g.,  PyTorch~\cite{paszke2019pytorch}).

\paragraph{Optimization process.} In Figure \ref{fig:opt_steps} we show the evolution of the triangulation through the optimization steps. The gradual change shows that indeed our differential triangulation enables gradient-based optimization which smoothly decreases the energy towards a local minimum. Please refer to the supplemental video for more detailed visualizations of the optimization process. 

\subsection{Loss blending}
\label{sec:loss_blending}
\begin{figure}
    \centering
    \includegraphics[width=0.8\linewidth]{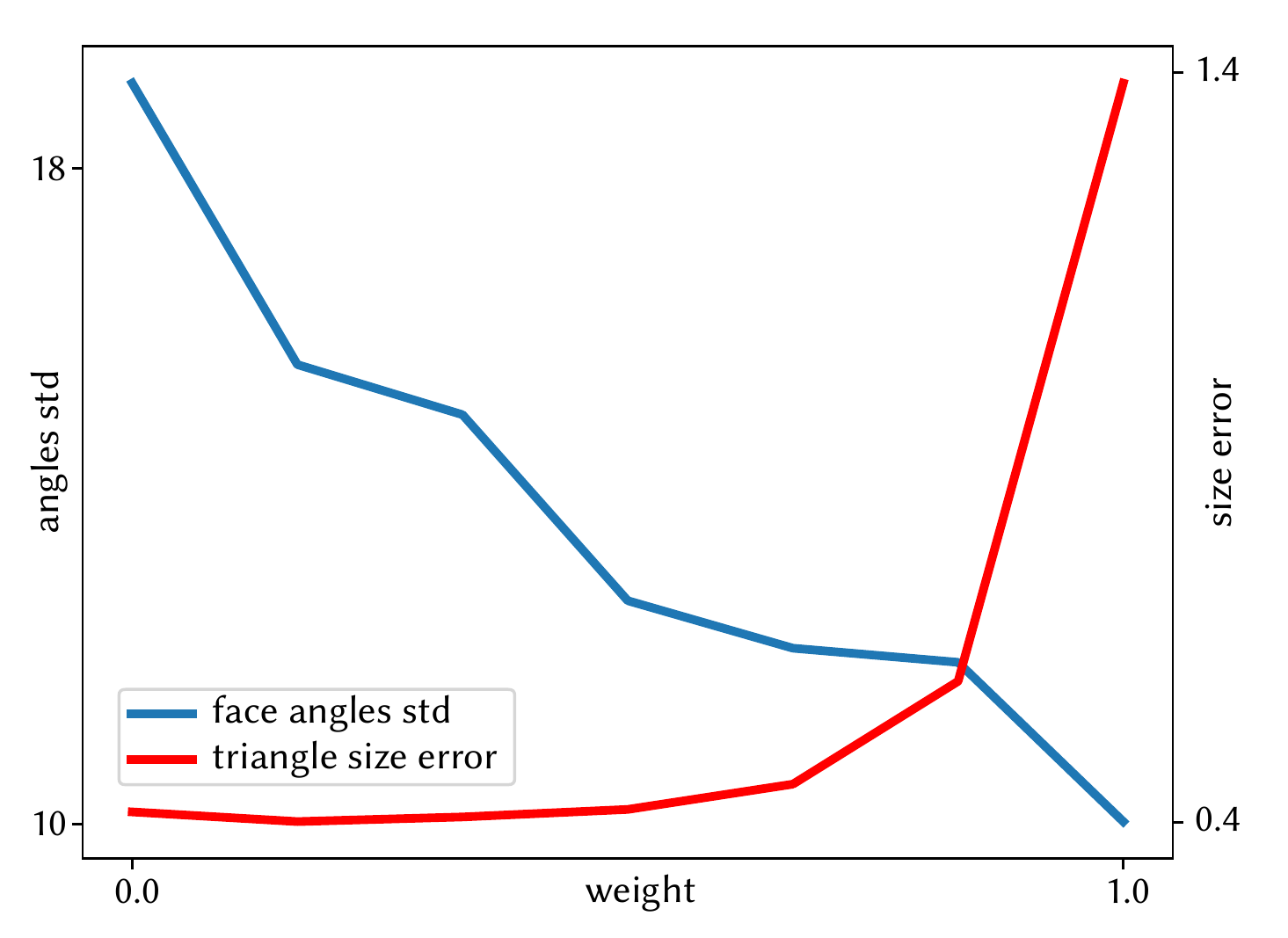}
    \caption{\textbf{Loss blending.} We blend the triangle sizing loss  $\mathcal{L}_s$ and the equilateral triangle loss $\mathcal{L}_a$ on 7 models of the dataset. We show the error in triangle distribution and the standard deviation of face angles. Note that given a triangulation, the average angle value is  60$^{\circ}$.  We observe that we can combine the two losses to obtain a trade off between the desired properties. }
    \label{fig:loss_blending}
\end{figure}

As an important advantage, our method naturally enables blending and interpolating the relative weights placed on different loss terms, such as triangle size and adherence to equilateral triangles. We show the plot of energies with respect to such a blending in Figure \ref{fig:loss_blending} using the aggregated loss term defined as, 
\begin{equation}
   \mathcal{L}(V', P) := t \times \mathcal{L}_a + (1-t) \times \mathcal{L}_s
  \label{eq:blending_loss}
\end{equation}
with $t$ being the blending weight. We evaluate over 7 values of the weight on a subset of 7 models from our dataset. This allows us to easily trade off between characteristics for the triangulation. Note that this was not previously possible for specialized methods targeted towards individual tasks.

\subsection{Method of vertex initialization}
We compare our method with an alternative vertex initialization technique in Table~\ref{tab:vertex_initizalization}. Given fixed per-patch boundary vertices, we uniformly sample the remaining vertices on the 3D surface using rejection sampling. We evaluate both initialization methods on the same subset of shapes from Section \ref{sec:loss_blending}. We observe that the alternative initialization method produces initial triangulations with higher errors. While our method is not completely insensitive to the initialization strategy, it can significantly decrease the loss in both cases.

\begin{table}%
\caption{\textbf{Vertex initialization methods.} We compare an initialization based on remeshed 3D vertices to an initialization based on a uniform distribution over the 3D surface. The uniform initialization has significantly higher initial error, but our method can still decrease the loss significantly.}
\label{tab:vertex_initizalization}
\begin{minipage}{\columnwidth}
\begin{center}
\begin{tabularx}{\linewidth}{lll}
  \toprule
     init. method & input mesh & ours  \\ \midrule
     remeshed model vertices   &1.354 & 0.634   \\
     uniform  &1.429& 0.791   \\
  \bottomrule
\end{tabularx}
\subcaption*{triangle size}
\begin{tabularx}{\linewidth}{lll}
  \toprule
     init. method &  input mesh & ours  \\ \midrule
     remeshed model vertices  &13.809 & 9.037   \\
     uniform  &19.119 & 11.028   \\
  \bottomrule
\end{tabularx}
\subcaption*{curvature alignment}
\end{center}
\end{minipage}
\end{table}%




\subsection{Runtime and memory}
\change{We show the average runtime and maximum memory usage of our method for multiple values of k in Table~\ref{tab:runtime_and_memory_k} and multiple vertices count per patch in Table \ref{tab:runtime_and_memory_vertex}.} The runtime is for a typical optimization with 1000 iterations. Both time and memory are linear in the number of vertices and cubic in the number of neighbors k. \change{In our experiments, we typically optimize for 1000 steps for the curvature alignment task   and  1500 steps for  the triangle size task.}

\begin{table}%
\caption{\textbf{Runtime and memory usage w.r.t. k.} We report the average runtime for an optimization with 1K iterations and maximum memory usage per patch.}
\label{tab:runtime_and_memory_k}
\begin{center}
\begin{tabular}{llll}
  \toprule
     k & 70 & 80 &90  \\ \midrule
     runtime  (sec) &132 & 185 &252   \\
     memory (GB)  &7.6&9.47&13.2   \\
  \bottomrule
\end{tabular}
\end{center}
\end{table}%

\begin{table}%
\caption{\change{\textbf{Runtime and memory usage w.r.t. number of vertices per patch.} We report the average runtime for an optimization with 1K iterations and maximum memory usage per patches of varying number of vertices. Time and memory are linear in the number of vertices. Note that we can adjust the size of our patches as needed to avoid memory limitations.} }
\label{tab:runtime_and_memory_vertex}
\begin{center}
\begin{tabular}{lllll}
  \toprule
     n vertices & 500 & 700 &900 & 1100  \\ \midrule
     runtime  (sec) &266 & 364 &477 &581   \\
     memory (GB)  &9.0&12.6&16.3 &21.1   \\
  \bottomrule
\end{tabular}
\end{center}
\end{table}%

\subsection{Analytic Surfaces}
\begin{figure}[t]
    \centering
    \includegraphics[width=\linewidth]{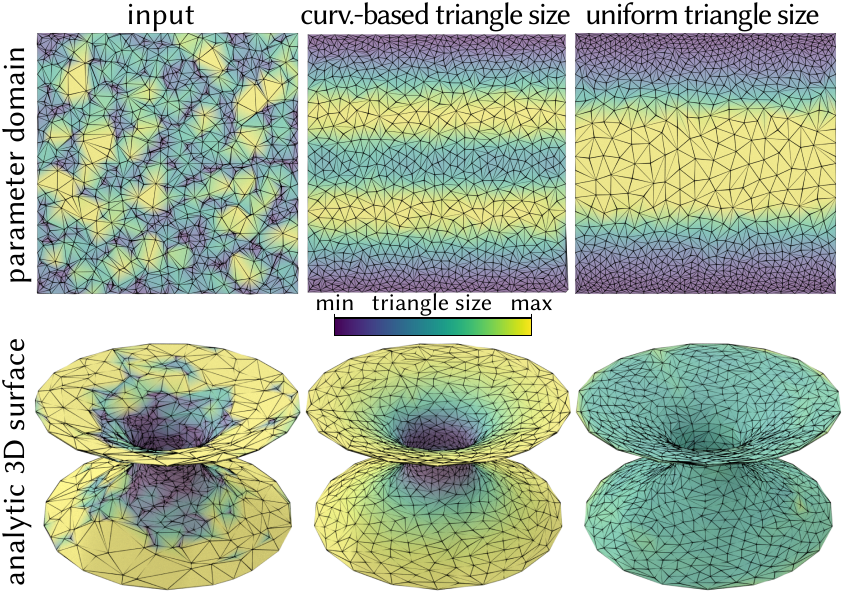}
    \caption{\textbf{Analytic surfaces.} Our method can triangulate surfaces given in any representation: here we triangulate an analytic surface (a catenoid), with the parameteric domain shown on the top row, and the 3D surface on the bottom row. Starting from randomly distributed vertices (left), our approach successfully triangulates the analytic surface with curvature-based triangle sizes (middle) and equal triangle sizes (right).}
    \label{fig:analytic}
    \vspace{-10pt}
\end{figure}

Our differentiable triangulation method can be applied to any kind of 3D surface, as long as bijective piecewise differentiable parameterization of the surface is available. In Figure~\ref{fig:analytic}, we experiment with an analytically defined 3D surface, a catenoid~\cite{dierkes2010minimal}. This surface is defined as a function over a 2D parameter domain 
(thus, the analytical function itself is our mapping $m$). We start with randomly distributed vertices in the parameter domain and optimize for either triangle sizes based on the curvature magnitude that we compute analytically or for equal-sized triangles in the 3D domain. Curvature values were computed analytically from the surface definition. We can see that our approach successfully optimizes these objectives on the analytic surface.

\subsection{\change{
Discussion on performance}}

\change{
We observe that our method presents an overhead compared to task-specific methods in terms of running time and the size of processed meshes. But with this overhead, our method buys generality and differentiability. We can minimize multiple different objectives (like the objectives of \cite{loseille2017unstructured}, \cite{jakob2015instant}), or can easily combine multiple objectives, without modifying our pipeline and our method can be used as a component in a differentiable framework. Our numerical results are on par with specifically tailored remeshing methods.
Note that several recent learning-based methods work with even smaller point counts (1024 points in PointNet \cite{qi2017pointnet}, 2250 Edges in MeshCNN \cite{hanocka2019meshcnn}, 2048 points in \cite{luo2021diffusion}), but these restrictions are quickly decreasing with improvements in GPU hardware and methodological improvements.
}

%% file: sections/conclusion.tex
\section{Conclusion, Limitations \& Future Work}
The framework presented in this paper is the first, to the best of our knowledge, to enable approaching surface triangulation from a differentiable point of view. As shown in the experiments, differentiability enables a generic and flexible framework, which can handle various geometric losses, along with their combinations, while taking advantage of modern optimization frameworks. We believe it is the first step towards a black-box, differentiable triangulation module in  deep learning frameworks such as PyTorch and TensorFlow where it can be immensely helpful in devising a trainable pipeline, e.g., learning to triangulate models based on deformation sequences.

Our method has two main limitations, the first of which is that the surface needs to be segmented into patches before triangulating. The boundaries of these patches do not participate in the optimization and hence some visible artifacts exist across boundaries. Nevertheless, we note that as shown in the experiments, even with this limitation, our approach achieves significantly better results than the state of the art. A possible solution for this would be to repeat the meshing by iteratively selecting different patches and reparameterizing until convergence. 

The second limitation of our method is that it cannot yet handle a large number of points (e.g., 100k+), or large patches, as we need to compute the \change{inclusion scores} over a the large space of possible triangles. As future work, we plan to consider a multiscale approach for tackling this issue. 

%

We are excited about the possibilities our approach opens up. For one, since our method can work with surfaces represented in any explicit format (as we show in Figure \ref{fig:analytic}), we wish to explore triangulating surfaces in other representations, such as NURBS, or neural representations such as AtlasNet~\cite{groueix2018papier,morreale2021neural}. Extending our method further to point clouds and recovering not only an optimized triangulation but also the topological structure \change{(i.e. the connectivity that defines a surface)} could be immensely important for future applications. As an immediate application, we wish to harness differentiabilty to \emph{train} a network to directly output vertex weights and displacements for any given surface in a single forward pass, and thus avoid test-time optimization.

\section{Acknowledgments}

Parts of this work were supported by the KAUST OSR Award No. CRG-2017-3426, the ERC Starting Grant No. 758800 (EXPROTEA), the ANR AI Chair AIGRETTE, gifts from Adobe, the UCL AI Centre and the European Union’s Horizon 2020 research and innovation programme under the Marie Skłodowska-Curie grant PRIME (No 956585).